\pdfoutput=1   
\documentclass[10pt,twocolumn,letterpaper]{article}

\usepackage[pagenumbers]{wacv}            

\usepackage{array}
\usepackage{colortbl}  
\usepackage[ruled,vlined]{algorithm2e}
\usepackage{balance}    

\definecolor{wacvblue}{rgb}{0.21,0.49,0.74}
\usepackage[pagebackref,breaklinks,colorlinks,allcolors=wacvblue]{hyperref}
\hypersetup{pdftitle={CRAD: Class-wise Reliability-Aware Distillation for Decentralized Heterogeneous Federated Learning},
            pdfauthor={Baraa Bilbeisi, Mengchen Fan, Baocheng Geng, Qing Tian}}
\usepackage{multirow}

\title{CRAD: Class-wise Reliability-Aware Distillation for Decentralized Heterogeneous Federated Learning}

\author{Baraa Bilbeisi$^{1,*}$\quad Mengchen Fan$^{2,*}$\quad Baocheng Geng$^{2}$\quad Qing Tian$^{2,\dagger}$\\[2pt]
$^{1}$Dept.\ of Electrical \& Computer Engineering \quad
$^{2}$Dept.\ of Computer Science\\
University of Alabama at Birmingham, AL, USA
}

\begin{document}
\maketitle
\iftoggle{wacvfinal}{%
  \renewcommand{\thefootnote}{\fnsymbol{footnote}}%
  \footnotetext[1]{These authors contributed equally.}%
  \footnotetext[2]{Corresponding author: \texttt{qtian@uab.edu}}%
  \renewcommand{\thefootnote}{\arabic{footnote}}%
}{}

\begin{abstract}
Conventional federated learning (FL) relies on parameter averaging, which forces clients to be doubly homogeneous: it demands an identical architecture and degrades under non-IID data. Real-world deployments usually break both assumptions. We sidestep both by building a decentralized knowledge distillation framework in which each client evaluates its peers' model snapshots on its own local data and distills from the resulting soft predictions. Because knowledge is transferred through the shared class posterior, clients are free to run different architectures; and because every teacher is evaluated on the student's own device, raw data never leaves the client, with no central server or public dataset required. Within this setting, we identify and address an under-examined problem: how to combine the peer teacher predictions. Existing methods, like uniform averaging, ignore how knowledge reliability varies across teachers and classes. We propose Class-wise Reliability-Aware Distillation (CRAD), which, per class, first discards teachers that disagree with the peer consensus and then takes a weighted average of the rest, weighting each teacher by its per-class reliability (precision, or inverse variance). Since the variance of an accuracy from $n$ samples scales as $1/n$, support enters automatically: among the teachers that survive filtering, a teacher is trusted for a class to the degree that it is both accurate and well-evidenced for it. On CIFAR-10, CIFAR-100, and PathMNIST, across heterogeneous architectures under severe non-IID skew, CRAD consistently outperforms competing methods in global accuracy. Our code is publicly available at \url{https://github.com/bbilbeis/CRAD}.
\end{abstract}

\section{Introduction}
\label{sec:intro}

Federated learning (FL) has enabled collaborative training of machine learning models across distributed clients while keeping raw data local \cite{McMahan2017AISTATS,McMahan2017Blog}. However, a fundamental limitation in practical deployments is the architectural homogeneity imposed by parameter averaging \cite{Fan2024Survey,Mora2024KDGuide,Khalil2024DFML}. 
This assumption is rarely satisfied in real-world FL systems \cite{Khalil2024DFML,Pfeiffer2023Survey,Zhang2025AdaptFL}.
Consider a network of hospitals collaborating to train a diagnostic imaging model. Each operates under distinct hardware and data availability: a large urban hospital may run a full-scale network on a GPU server, while a rural clinic can only afford a compact model on a low-power device.
Autonomous-driving fleets face the same problem: vehicles carry different onboard hardware and run in different environments. Because participants in either setting cannot share raw data, FL is the natural solution \cite{Kairouz2021FL,Asad2026FedProx}.
Yet standard FL fails here. Averaging across different architectures is undefined, since the weight tensors differ in shape and semantics \cite{Fan2024Survey,Lin2020NeurIPS}. Forcing a shared architecture is no remedy. No single architecture suits all institutions without over-provisioning weak devices or under-using capable ones \cite{Pfeiffer2023Survey,Zhang2025AdaptFL}, and averaging models that have each specialized to their own skewed data can produce a model worse than any individual specialist \cite{Zhao2018NonIID,Briggs2020IJCNN,Jimenez2025Assessment}.

Existing FL methods fall into two groups, neither fully suited to decentralized, architecturally heterogeneous learning. Parameter-space consensus methods \cite{Li2020MLSys,Karimireddy2020ICML,Wang2020NeurIPS} curb client drift under non-IID data but still combine weights within a shared model space, so they remain inapplicable across incompatible architectures \cite{Mora2024KDGuide,Khalil2024DFML}. Knowledge-distillation methods \cite{Lin2020NeurIPS,Itahara2023Distill,Li2019FedMD} instead exchange architecture-agnostic soft predictions, escaping the homogeneity constraint, but typically reintroduce a shared public dataset or a central server \cite{Mora2024KDGuide,Guo2025DataFreeKD}, the very privacy and centralization costs FL is meant to avoid \cite{Malinovsky2023Regularized,Nguyen2022DecentralizedMed}.
These observations motivate our approach: a fully decentralized, peer-to-peer knowledge distillation (KD) framework in which clients collaborate without parameter averaging. Each client shares model snapshots, instantiates peer snapshots as local teachers, evaluates them on its own data, and distills from their combined soft predictions, with no public transfer set or central coordinator. Decentralized, public-data-free distillation of this kind is attractive because it is natively compatible with heterogeneous architectures, and it lets each client learn the classes it sees rarely or not at all \cite{Khalil2024DFML}. But making this setting work surfaces a problem the FL distillation literature has largely left implicit: once a client holds a pool of peer teacher predictions, how should those predictions be combined to guide the student?

The default answer, inherited from ensemble distillation \cite{Lin2020NeurIPS}, is a uniform average over all peers. This is a poor choice under heterogeneity, where teacher reliability is uneven and class-dependent. The central question of our work is therefore how a client should weight its peer teachers, per class, accounting for both how accurate and how well-supported each teacher is (a class accuracy from few samples being itself unreliable), without a server, public data, or expensive tuning.

We answer with Class-wise Reliability-Aware Distillation (CRAD). Alongside its model snapshot, each client shares a compact class-wise statistics vector summarizing, for every class, how many training and validation samples it holds and how often it classifies the latter correctly. On receiving these statistics from its peers, a client works class by class: it first discards the teachers that hold insufficient samples of the class or whose predictions deviate most from the per-class peer consensus, then combines the survivors by inverse-variance (precision) weighting, so that each teacher counts in proportion to both the decisiveness and the statistical support of its class-accuracy estimate. Because the variance of an accuracy estimated from $n$ samples scales as $1/n$, sample support enters the weight automatically. A teacher that survives the filter and is both accurate and well-supported for a class receives a large weight, while a teacher guessing from a few samples is discounted. Our contributions are:

\begin{itemize}
    \item We present a decentralized, public-data-free FL framework that handles model and data heterogeneity at once: clients of differing architectures collaborate purely through peer distillation over the shared $C$-dimensional class posterior. Within it, we surface an under-examined problem of how to combine peer teacher predictions, and show that, since teacher reliability is uneven across peers and classes, the usual uniform averaging is a poor choice.
    \item As our central methodological contribution, we propose CRAD, which, per class, filters out teachers that disagree with the peer consensus and then weights the survivors by the precision (inverse variance) of their class-accuracy estimates. Up to smoothing constants, this precision equals the Fisher information of each estimate, the amount of statistical evidence behind the teacher's competence for that class. It rewards both accuracy and sample support, requires few tuned hyperparameters, and adds only a compact per-class statistics vector to communication.
    \item On the CIFAR-10, CIFAR-100, and PathMNIST benchmarks, under heterogeneous architectures and non-IID skew, CRAD attains the best global accuracy of all compared methods while maintaining strong local accuracy.
\end{itemize}

\begin{figure*}[t]
    \centering
    \includegraphics[width=0.95\textwidth]{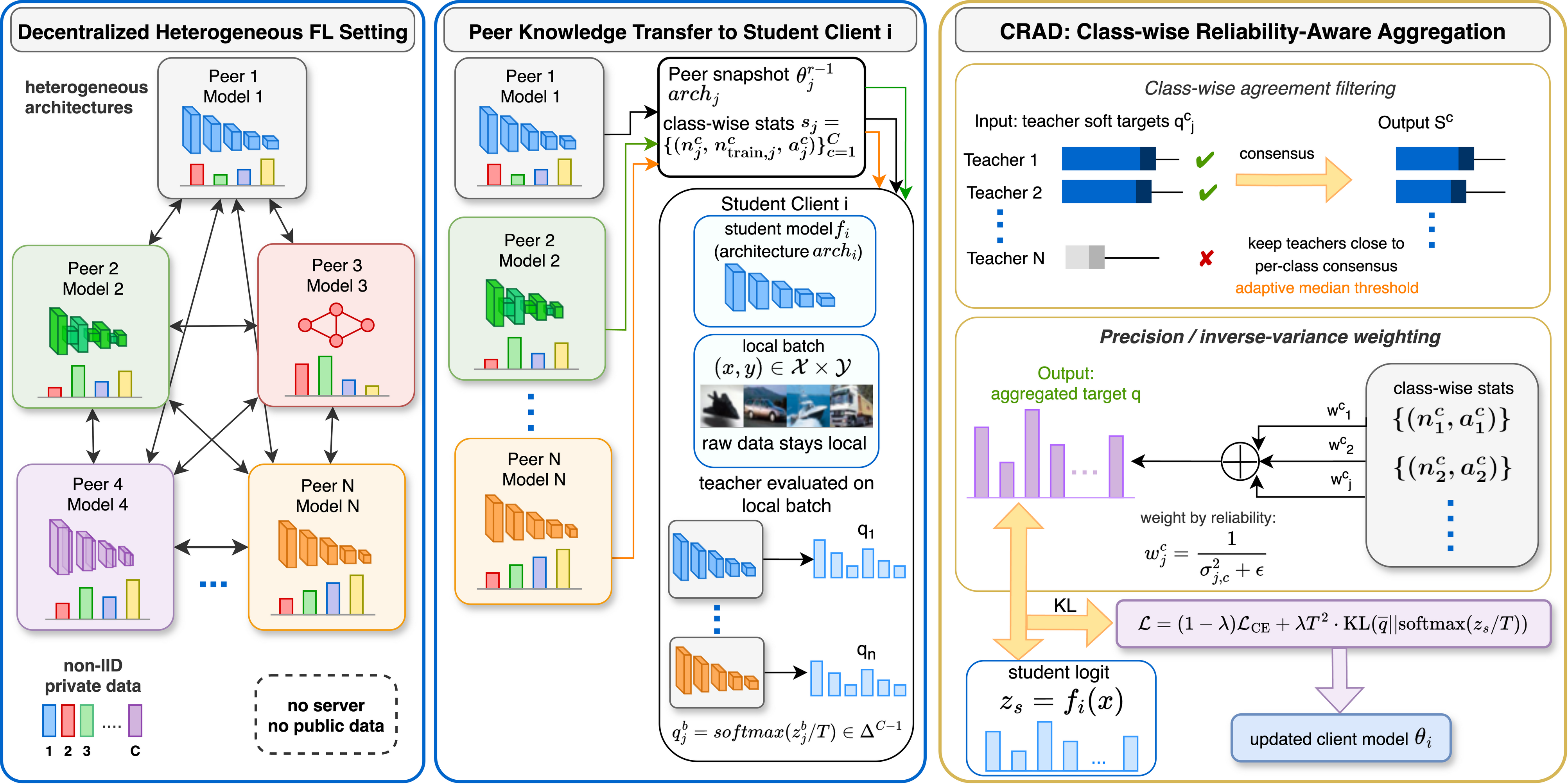}
    \caption{
    Overview of the proposed CRAD framework.
    \textbf{(1)~Decentralized heterogeneous setting:} clients running different architectures collaborate over non-IID private data, with no central server and no public dataset. The only shared interface between any two clients is the $C$-dimensional class posterior.
    \textbf{(2)~Peer knowledge transfer:} a student client receives its peers' model snapshots together with a compact per-class statistics vector, and runs each peer's snapshot on its own local batch to obtain soft predictions $q_j$, so raw and private data never leaves the client.
    \textbf{(3)~Class-wise reliability-aware aggregation:} for each class, CRAD first discards the teachers that disagree with the peer consensus (an adaptive per-class median filter), then combines the survivors by precision (inverse-variance) weighting of their class-accuracy estimates, so that decisive, well-supported teachers dominate the aggregated distillation target $\overline{q}$.
    The student then updates its model by minimizing a combination of cross-entropy on its local labels and a distillation loss toward this target.
    }
    \label{fig:crad_overview}
\end{figure*}

\section{Related Work}
\label{sec:relwork}

\subsection{Data Heterogeneity in Federated Learning}

FL degrades under data heterogeneity, the non-IID data problem \cite{Zhao2018NonIID}. Previous efforts to mitigate this degradation have focused on parameter-space regularization to bound client drift, such as FedProx \cite{Li2020MLSys}, or variance reduction via control variates, such as SCAFFOLD \cite{Karimireddy2020ICML}, to stabilize convergence. Another series of works emphasizes model personalization, allowing clients to adapt to local distributions through partial parameter sharing, meta-learning, or personalized objective functions, such as pFedMe \cite{Dinh2020pFedMe} and APFL \cite{Deng2020APFL}. However, all of these operate within a shared parameter space, whether by global, proximal, or personalized aggregation \cite{Fan2024Survey}, and so remain incompatible with architecturally distinct models \cite{Fan2024Survey,Pfeiffer2023Survey}, a setting common in practice.

\subsection{Knowledge Distillation in Federated Learning}

KD sidesteps the homogeneity constraint by transferring knowledge through soft predictions rather than by averaging weight vectors. Predictions stay comparable across architectures over a shared label space \cite{Mora2024KDGuide,Salman2025Survey}. To fuse those predictions into a common target, KD-based federated methods fall into three groups, distinguished by the auxiliary infrastructure they require. The first group aligns clients on a shared public dataset without a central server, distilling each client toward a consensus formed on the common inputs, as in FedMD \cite{Li2019FedMD}. This transfers knowledge across architectures but requires a task-relevant public dataset, raising availability and relevance concerns \cite{Guo2025DataFreeKD}. The second group routes fusion through a central server that distills client predictions into a global consensus and broadcasts it back, as in ensemble distillation \cite{Lin2020NeurIPS}. This typically runs on the server's own public or generated data, relocating the dependency rather than removing it, and reintroduces a central coordinator \cite{Khalil2024DFML,Nguyen2022DecentralizedMed}. Recent server-centric variants follow the same pattern. The multi-teacher FedMKD of Sun et al.~\cite{Sun2025FedMKD} distills client models into a server-side student with one confidence scalar per teacher, and the personalized FedMKD of Lin et al.~\cite{Lin2025FedMKD} initializes each client by blending the global model with the client's own previous-round model in parameter space. Both require a central server and architecturally identical clients. A related self-distillation line~\cite{Lee2022FedNTD,He2022FedSSD} selectively distills the global model into the local one, filtering or reweighting its knowledge class by class (and, in~\cite{He2022FedSSD}, sample by sample), but operates within FedAvg, with a single server-aggregated teacher and identical client architectures. The third group needs neither a public dataset nor a central server. Clients distill from one another peer-to-peer over their own data, as in DFML \cite{Khalil2024DFML} and personalized decentralized KD \cite{Jeong2023DecentralizedKD}. Our framework operates in this regime.

\subsection{Combining Teacher Predictions}

In most KD-based federated methods, peer predictions are fused by a uniform average over the available teachers \cite{Lin2020NeurIPS,Li2019FedMD}. This treats every teacher as equally trustworthy, but teachers trained on different data partitions vary in quality, and their reliability is itself class-dependent. Under non-IID partitions, a given peer may be reliable on its well-represented classes and badly miscalibrated elsewhere. A natural alternative, used in centralized multi-teacher distillation \cite{Zhang2022CAMKD}, is to weight teachers by prediction confidence or entropy, but confidence is a poor proxy for correctness under distribution shift, and such schemes ignore how much data actually supports each teacher's belief. Reliability-based weighting normally assumes held-out validation data on a trusted server or access to a shared transfer set, neither of which exists in our setting. How to weight teachers by reliability in a fully decentralized, public-data-free, non-IID setting, accounting for the statistical uncertainty of reliability estimated from finite local data, remains largely open. Our work targets this gap. We keep the decentralized, public-data-free KD backbone and present a per-class aggregation that first filters teachers by agreement and then weights each survivor by the precision of its class-accuracy estimate.

\section{Method}
\label{sec:method}

\subsection{Setting and Overview}

We consider $N$ clients, each holding a private local dataset $\mathcal{D}_i$ of input-label pairs $(x,y) \in \mathcal{X} \times \mathcal{Y}$ drawn from a client-specific distribution $P_i$, with a label space $\mathcal{Y}=\{1,\ldots,C\}$ shared across clients. The federation is heterogeneous in two ways at once. The data are non-IID: the class proportions $P_i(y)$ differ sharply across clients, and under severe skew a client may hold few or even no samples of some classes, so its local label set can be a strict subset of $\mathcal{Y}$. The models are heterogeneous: each client $i$ runs a network $f_i$ of its own architecture, scaled to its hardware. These are exactly the conditions under which standard Federated Averaging breaks down: weight averaging is undefined across heterogeneous architectures, and even within one, averaging classifiers trained on skewed partitions can degrade the global model below a local specialist \cite{Zhao2018NonIID,Fallah2020NeurIPS,Li2020MLSys}.

We therefore discard parameter averaging in favor of fully decentralized, peer-to-peer knowledge distillation, and within it focus on how peer predictions are fused into a distillation target. The backbone departs from parameter averaging across three coupled dimensions. Rather than averaging in parameter space, knowledge is transferred through soft output distributions produced by teacher models evaluated locally on each client's own data. Rather than requiring architectural uniformity, each client maintains a structurally distinct model. The only interface between teacher and student is the semantically uniform $C$-dimensional class posterior. Rather than relying on a central server, we operate over a fully connected peer-to-peer topology in which every client communicates directly with all others, with no global coordinator at any stage of training. On top of this backbone, CRAD forms the distillation target through a per-class procedure in two stages: it first discards teachers that disagree with the per-class peer consensus, then combines the survivors with each teacher's contribution scaled by the statistical precision of its class-accuracy estimate. The consensus the student aligns with is thus dominated, for every class, by the teachers that are both in agreement and well-evidenced for it.

\subsection{Architecture-Agnostic Knowledge Transfer}

At communication round $r$, each client $i$ maintains its own model $f_i$ and receives a set of model snapshots $\{(\theta_j^{r-1},\mathrm{arch}_j,\mathbf{s}_j^{r-1})\}_{j\in\mathcal{N}_i}$ from its peers, where $\mathcal{N}_i=\{1,\ldots,N\}\setminus\{i\}$ denotes the set of all other clients, $\mathrm{arch}_j$ denotes the architecture identifier for client $j$, and $\mathbf{s}_j^{r-1}$ is a compact class-wise statistics vector defined in Section~\ref{sec:crad}. Notably, all clients in round $r$ distill from round $r-1$ snapshots.
For each training batch $(x^{(b)},y^{(b)})$ sampled from $\mathcal{D}_i$, client $i$ instantiates each neighbor's architecture $f_{\theta_j^{r-1}}$ locally and computes the teacher logits:
\begin{equation}
z_j^{(b)} = f_{\theta_j^{r-1}}(x^{(b)}),\quad j\in\mathcal{N}_i,
\label{eq:teacher_logits}
\end{equation}
followed by a temperature-scaled softmax ($T>0$) that softens the distribution \cite{Hinton2015KD}, yielding the soft target vector $q_j^{(b)}$:
\begin{equation}
q_j^{(b)} = \mathrm{softmax}(z_j^{(b)}/T) \in \Delta^{C-1},
\label{eq:teacher_soft_target}
\end{equation}
where $\Delta^{C-1}$ denotes the probability simplex over $C$ classes. The soft target $q_j^{(b)}$ encodes what neighbor $j$ believes about the class structure of the student's own local sample $x^{(b)}$, including classes the student may have never encountered. This is the mechanism by which knowledge of unseen classes propagates across the network without a shared dataset or central coordinator. Because teacher inference runs entirely on the student's data, no raw samples leave the client. Because each teacher functions as a black-box mapping $x\rightarrow\mathbb{R}^{C}$, any two models sharing the same label space $\mathcal{Y}$ are mutually compatible for knowledge transfer regardless of their internal architecture.

\subsection{Class-wise Reliability-Aware Aggregation}
\label{sec:crad}

Given the per-teacher soft targets $\{q_j^{(b)}\}_{j\in\mathcal{N}_i}$ of Eq.~\ref{eq:teacher_soft_target}, the remaining question is how to combine them into a single distillation target. The standard choice is the uniform mean over all peers \cite{Lin2020NeurIPS},
\begin{equation}
\overline{q}^{(b)}_{\mathrm{unif}} = \frac{1}{|\mathcal{N}_i|}\sum_{j\in\mathcal{N}_i} q_j^{(b)},
\label{eq:prelim_consensus}
\end{equation}
which weights every teacher equally for every class. This is wasteful under heterogeneity. A teacher's competence is class-dependent, likely high for classes it has seen often and low for classes it has seen rarely or never. We instead operate separately for each class in two stages: we first discard teachers that lack support for the class or disagree with the per-class peer consensus, then aggregate the survivors weighted by an estimate of their reliability for that class.

\paragraph{Class-wise agreement filtering.}
Before weighting, we first drop, for each class, teachers whose local training class support falls below a per-class threshold $n_{\min}^c$, set to a quantile of the peers' training support counts $\{n_{\mathrm{train},j}^c\}_{j \in \mathcal{N}_i}$ (see Table~\ref{tab:main} caption), defining a competent teacher subset $\mathcal{N}_i^{c} = \{\, j \in \mathcal{N}_i : n_{\mathrm{train},j}^c \ge n_{\min}^c \,\}$. Among these remaining teachers, we then discard those whose predictions deviate most from the peer consensus on that class. To this end, we form a precision-weighted preliminary per-class consensus and measure each teacher's deviation from it,
\begin{equation}
\tilde{q}^{c} = \frac{\sum_{j\in\mathcal{N}_i^{c}} w_j^c\, q_j^{c}}{\sum_{j\in\mathcal{N}_i^{c}} w_j^c + \epsilon},
\qquad
d_j^{c} = \left|\, q_j^{c} - \tilde{q}^{c} \,\right|,
\label{eq:agreement_score}
\end{equation}
where $w_j^c$ is the precision weight defined below in Eq.~\ref{eq:precision_weight}, and retain for each class the teachers whose deviation is at most the median deviation for that class,
\begin{equation}
\tau_c = \operatorname*{median}_{j\in\mathcal{N}_i^{c}} d_j^{c},
\qquad
\mathcal{S}_i^{c} = \left\{\, j\in\mathcal{N}_i^{c} \;:\; d_j^{c} \le \tau_c \,\right\}.
\label{eq:selection}
\end{equation}
This adaptive, per-class median retains the supported teachers most in agreement with the consensus and introduces no threshold to tune. The class-wise aggregation below then ranges over $\mathcal{S}_i^{c}$ rather than the full peer set, and different classes may draw on different teacher subsets.

\paragraph{Class-wise reliability statistics.}
Alongside its snapshot, each client $j$ shares a compact statistics vector $\mathbf{s}_j=\{(n_j^c, n_{\mathrm{train},j}^c, a_j^c)\}_{c=1}^{C}$ recording, for each class $c$, its class-$c$ accuracy $a_j^c$, evaluated with its current model on a held-out validation split of its own shard, the number $n_j^c$ of class-$c$ examples in that split, on which the accuracy is computed, and the number $n_{\mathrm{train},j}^c$ of class-$c$ training examples, which determines the competent teacher subset $\mathcal{N}_i^c$. This is $3C$ scalars per client per round, negligible compared to the model weights, and exposes no raw samples.

\paragraph{Reliability as a statistical estimate.}
The accuracy $a_j^c$ is a noisy signal: the same value is more credible when measured on many class-$c$ examples than on a handful. We therefore treat $a_j^c$ as a Bernoulli proportion over the $n_j^c$ examples it was measured on, applying the Agresti--Coull correction to stabilize small-sample estimates and taking its associated variance,
\begin{equation}
\tilde{a}_j^c = \frac{a_j^c\, n_j^c + 2}{n_j^c + 4},
\qquad
\sigma_{j,c}^2 = \frac{\tilde{a}_j^c\,(1-\tilde{a}_j^c)}{n_j^c + 4}.
\label{eq:variance}
\end{equation}

\paragraph{Inverse-variance (precision) weighting.}

We weight each surviving teacher by the precision (inverse variance) of its class-accuracy estimate. The weight rises with both the decisiveness of the estimated accuracy and the amount of evidence behind it (up to the Agresti--Coull smoothing constants, the precision equals the Fisher information of the estimate). Precision is the standard rule for combining estimates of differing reliability \cite{Cochran1954}. Because precision accrues linearly with additional samples, a teacher's influence grows in proportion to its evidence. Weights not proportional to $1/\sigma^2$ break this proportionality, and every alternative we tested performs worse (Tab.~\ref{tab:cifar100_weight_ablation}). We thus set the class-wise weight
\begin{equation}
w_j^c = \frac{1}{\sigma_{j,c}^2 + \epsilon},
\label{eq:precision_weight}
\end{equation}
where $\epsilon>0$ guards against division by zero. We aggregate the surviving teachers class by class as a precision-weighted average:
\begin{equation}
\overline{q}^{c} = \frac{\sum_{j\in\mathcal{S}_i^{c}} w_j^c\, q_j^{c}}{\sum_{j\in\mathcal{S}_i^{c}} w_j^c + \epsilon},
\qquad
\overline{q} \leftarrow \frac{\overline{q}}{\sum_{c=1}^{C}\overline{q}^{c}},
\label{eq:crad_aggregate}
\end{equation}
where $\mathcal{S}_i^{c}$ is the agreement-filtered teacher set of Eq.~\ref{eq:selection} and the renormalization restores $\overline{q}$ to the probability simplex (we omit the batch index $b$ on $\overline{q}^c$, $q_j^c$ for brevity). The weight is a per-class trust score on the teacher's predictions. Precision alone rewards decisiveness rather than correctness, so the weighting is applied only after the class-wise filter removes teachers with insufficient support or predictions that substantially deviate from the peer consensus. Under the standard assumption that unreliable or outlying teachers constitute a minority, this filtering reduces the risk that a confidently inaccurate teacher receives a high weight. A teacher is trusted for a class to the degree that it is confident, well-evidenced, and in consensus. That said, repeated mutual distillation couples the teachers across rounds, so their predictions are not independent. We use the weight as a trust score rather than as a minimum-variance combiner, for which independence would be required.

Overall, the training objective balances local cross-entropy supervision, $\mathcal{L}_{\mathrm{CE}}$, with a distillation term that aligns the student's predictions with the precision-weighted consensus:
\begin{equation}
\mathcal{L}=(1-\lambda)\,\mathcal{L}_{\mathrm{CE}}+\lambda\,T^2\cdot\mathrm{KL}\left(\overline{q}\;\|\;\mathrm{softmax}(z_s/T)\right),
\label{eq:total_loss}
\end{equation}
where $z_s=f_i(x^{(b)})$ is the student's own logit vector and $\lambda\in[0,1]$ is the distillation weight, balancing reliance on the peer consensus against the local ground-truth labels. The temperature-scaled KL term with its $T^2$ factor is the standard distillation objective, here applied to the aggregated target $\overline{q}$ rather than to a single teacher.
Algorithm~\ref{alg:crad} summarizes the complete CRAD framework.

\SetKwComment{Comment}{/* }{ */}
\SetKwInOut{KwIn}{Input}
\SetKwInOut{KwOut}{Output}
\SetAlgoCaptionLayout{normalsize}
\begin{algorithm*}[t]
\caption{Class-wise Reliability-Aware Distillation (CRAD)}
\label{alg:crad}
\normalsize

\KwIn{$N$ clients with local datasets $\{\mathcal{D}_i\}_{i=1}^N$, temperature $T$, loss weight $\lambda$, smoothing constant $\epsilon$}
\KwOut{Updated client models $\{f_{\theta_i}\}_{i=1}^N$}

Initialize model $f_{\theta_i}$ for all clients $i \in \{1,\ldots,N\}$\;

\For{each communication round $r = 0$ to $R-1$}{
    Freeze current models as snapshots: $\theta_{\mathrm{prev},i} \leftarrow \theta_i$ for all $i$\;
    Each client computes class statistics $\mathbf{s}_i=\{(n_{i}^c, n_{\mathrm{train},i}^c, a_i^c)\}_{c=1}^C$ (validation count $n_i^c$, training count $n_{\mathrm{train},i}^c$, and validation accuracy $a_i^c$)\;

    \For{each client $i = 1$ to $N$}{
        Define peer set $\mathcal{N}_i = \{1,\ldots,N\} \setminus \{i\}$ and, per class, the support-filtered subset $\mathcal{N}_i^c \subseteq \mathcal{N}_i$ of teachers with sufficient class-$c$ samples\;
        Compute precision weights from peer statistics:
        $\tilde{a}_j^c = \frac{a_j^c n_j^c+2}{n_j^c+4}$,\,
        $\sigma_{j,c}^2 = \frac{\tilde{a}_j^c(1-\tilde{a}_j^c)}{n_j^c+4}$,\,
        $w_j^c = \frac{1}{\sigma_{j,c}^2+\epsilon}$,\, $\forall j\in\mathcal{N}_i, c$\;

        \For{each batch $(x,y) \in \mathcal{D}_i$}{
            Student logits and cross-entropy loss:
            $z_s = f_{\theta_i}(x)$,\, $\mathcal{L}_{\mathrm{CE}} = \mathrm{CE}(z_s, y)$\;

            Compute teacher soft targets (no gradients):
            $q_j = \mathrm{Softmax}(f_{\theta_{\mathrm{prev},j}}(x)/T), \quad \forall j \in \mathcal{N}_i$\;

            Class-wise agreement filter (adaptive per-class median), over $\mathcal{N}_i^c$:
            $\tilde{q}^{c} = \frac{\sum_{j\in\mathcal{N}_i^c} w_j^c q_j^{c}}{\sum_{j\in\mathcal{N}_i^c} w_j^c + \epsilon}$,\,
            $\tau^c = \operatorname{median}_{j\in\mathcal{N}_i^c} |q_j^c - \tilde{q}^c|$,\,
            $\mathcal{S}^c = \{ j\in\mathcal{N}_i^c : |q_j^c - \tilde{q}^c| \le \tau^c \}$\;

            Filtered precision-weighted class-wise aggregate:
            $\overline{q}^{c} = \frac{\sum_{j\in\mathcal{S}^c} w_j^c q_j^{c}}{\sum_{j\in\mathcal{S}^c} w_j^c + \epsilon}$, then normalize $\overline{q}\leftarrow \overline{q}/\sum_c \overline{q}^c$\;

            KD and total loss:
            $\mathcal{L}_{\mathrm{KD}} = T^2 \cdot \mathrm{KL}(\overline{q} \,\|\, \mathrm{Softmax}(z_s/T))$,\,
            $\mathcal{L} = (1-\lambda) \mathcal{L}_{\mathrm{CE}} + \lambda \mathcal{L}_{\mathrm{KD}}$\;

            Update $\theta_i$ via backpropagation (with gradient clipping)\;
        }
    }
}
\end{algorithm*}

\section{Experiments}
\label{sec:experiments}

\subsection{Experimental Setup}

\paragraph{Datasets and Partitioning.}
We evaluate the proposed framework on three benchmarks from distinct domains: CIFAR-10 and CIFAR-100 for natural-image classification, and PathMNIST \cite{Yang2023MedMNIST} for colon-pathology histology. PathMNIST comprises nine tissue classes from the MedMNIST collection and represents a realistic, privacy-sensitive medical setting. Each dataset is partitioned across $N=10$ clients (unless otherwise specified) under a Dirichlet distribution with concentration parameter $\alpha = 0.3$.
This induces severe label skew. These public benchmarks serve only to construct the simulated federation: each client trains on its own private shard alone, and the method uses no shared or public data.

\paragraph{Architecture Pool.}
To validate architectural heterogeneity, the framework uses three structurally distinct models: ResNet-18 ($\approx 11$M parameters), ResNet-18-Half ($\approx 2.8$M parameters, $0.5\times$ channel scaling), and a plain CNN ($\approx 0.8$M parameters, 2 conv + 2 FC). This pool evaluates both width heterogeneity, through the channel scaling between the ResNet variants, and family heterogeneity, through the absence of residual connections in the plain CNN. Architectures are assigned deterministically across the 10 clients via a round-robin schedule, denoted as $\mathrm{arch}_i=\mathrm{POOL}[i\bmod 3]$. This capacity heterogeneity, combined with the non-IID skew, makes teacher reliability uneven and class-dependent, reflecting common real-world deployment scenarios.

\paragraph{Compared Methods.}
All methods are evaluated under the same heterogeneous architecture pool, non-IID partition, and training schedule. We compare against five representative heterogeneous-FL distillation baselines that span the auxiliary-component taxonomy of Section~\ref{sec:relwork}. FedMD~\cite{Li2019FedMD} aligns clients on a shared public proxy dataset (size $500$) before local adaptation, directly testing the value of our public-data-free design. FedGD~\cite{zhang2023target} replaces the public proxy with a synthetic transfer set (size $500$) generated rather than collected, a data-free variant of the same public-data paradigm. MSFKD~\cite{Wang2023Knowledge} is a multi-teacher selective federated distillation method. DFML~\cite{Khalil2024DFML} is the closest baseline to our backbone, a fully decentralized, public-data-free, peer-to-peer mutual distillation method that, like ours, exchanges no raw data and uses no server, but fuses peers without class-wise reliability weighting. FedMKD~\cite{Lin2025FedMKD} is a recent multi-teacher method that reuses each client's own previous-round model as an additional teacher. Its implementation is not public, so we adapt it to our decentralized, architecture-heterogeneous setting.
Within our framework, the central comparison isolates the effect of teacher aggregation. 
We compare CRAD against two aggregation variants: Uniform KD, the conventional uniform average (Eq.~\ref{eq:prelim_consensus}); and Uncertainty KD, which weights teachers by the negative entropy of their predictions.

\paragraph{Training Protocol.}
Each client is trained for $R=300$ communication rounds, one local epoch per round, with the distillation weight $\lambda=0.7$ and a random seed of 1024.
Each client's held-out partition is split evenly into a validation half and a test half. The validation half supplies the reliability statistics of Section~\ref{sec:crad} and selects the checkpoint. The test half is touched only for final reporting. We report two metrics: global accuracy on each benchmark's standard test set, never used during training or model selection (our primary metric, measuring generalization), and local accuracy on each client's held-out test half (measuring local adaptation). Because a client can score high local accuracy simply by overfitting the classes it holds, local accuracy serves mainly as a guardrail, confirming that global-accuracy gains do not come at a large cost in local adaptation. For every method, both metrics are reported at the round with the highest average local validation accuracy.

\subsection{Experimental Results}
Table~\ref{tab:main} reports all methods on all three benchmarks under identical conditions.
\paragraph{CIFAR-10.}
The three aggregation variants are comparable on local accuracy, consistent with its guardrail role. They differ chiefly on global accuracy, which draws on borrowed knowledge of rarely-seen classes. There CRAD leads at $78.60\%$, ahead of Uncertainty KD ($77.41\%$) and the uniform average ($77.19\%$). Uncertainty KD barely improves on uniform averaging because confidence is a weak proxy for correctness under distribution shift, whereas CRAD's agreement filter keeps only teachers that align with the peer consensus, and its precision weight then favors those whose class beliefs are both decisive and well-supported.
All competing methods trail our framework in global accuracy. FedMKD~\cite{Lin2025FedMKD} is the strongest at $74.54\%$, still four points below CRAD. The public-data methods FedMD, MSFKD and FedGD are limited by the quality of their proxy or synthetic transfer set. DFML ($61.99\%$), the decentralized public-data-free method closest to our backbone, trails here, although it is the strongest competitor on CIFAR-100. That every aggregation variant in our framework exceeds every competing method isolates the contribution of the decentralized backbone, while the gap between CRAD and Uniform KD isolates that of reliability-aware aggregation.

\begin{table}[t]
\centering
\caption{Teacher-aggregation comparison on CIFAR-10, CIFAR-100, and PathMNIST ($N=10$, non-IID Dirichlet, $R=300$). All decentralized-KD methods share the same backbone and distillation weight $\lambda=0.7$, differing only in how peer teachers are combined. The checkpoint is selected on each client's validation split. CRAD's support threshold $n_{\min}^c$ is the per-class peer median of local training samples on CIFAR-10 and CIFAR-100, and the 20th percentile on PathMNIST. Accuracies are percentages, with global accuracy the primary metric. Best aggregation result in each column in \textbf{bold}.}
\label{tab:main}
\setlength{\tabcolsep}{2.5pt}
\scriptsize
\resizebox{\columnwidth}{!}{%
\begin{tabular}{lcccccc}
\toprule
& \multicolumn{2}{c}{CIFAR-10} & \multicolumn{2}{c}{CIFAR-100} & \multicolumn{2}{c}{PathMNIST} \\
\cmidrule(lr){2-3}\cmidrule(lr){4-5}\cmidrule(lr){6-7}
Method & Global & Local & Global & Local & Global & Local \\
\midrule
\multicolumn{7}{l}{\textit{Competing heterogeneous-FL methods}}\\
FedMD~\cite{Li2019FedMD}       & 71.69 & 86.96 & 36.23 & 55.85 & 86.58 & 88.04 \\
FedGD~\cite{zhang2023target}   & 67.87 & 86.48 & 30.50 & 50.80 & 82.86 & 86.93 \\
MSFKD~\cite{Wang2023Knowledge} & 69.22 & 86.44 & 32.77 & 50.63 & 62.19 & 72.04 \\
DFML~\cite{Khalil2024DFML}         & 61.99 & 66.88 & 47.29 & 54.12 & 70.40 & 74.06 \\
FedMKD~\cite{Lin2025FedMKD}    & 74.54 & 87.30 & 38.05 & 55.59 & 88.35 & 87.65 \\
\midrule
\multicolumn{7}{l}{\textit{Our framework, varying teacher aggregation}}\\
Uniform KD                     & 77.19& 87.73& 41.53& 58.54& 87.59& 87.89\\
Uncertainty KD & 77.41& \textbf{87.87}& 42.06& 58.74& 86.41& 87.09\\
\rowcolor{gray!20} \textbf{CRAD (Ours)} & \textbf{78.60} & 87.09 & \textbf{48.22} & \textbf{59.87} & \textbf{89.44} & \textbf{88.14}\\
\bottomrule
\end{tabular}}
\end{table}

\paragraph{Scaling to More Classes (CIFAR-100).}
The CIFAR-100 columns of Table~\ref{tab:main} cover a 100-class benchmark in which each client holds far fewer samples per class, stressing the reliability estimates that drive CRAD.
On this harder benchmark, CRAD again ranks first, at $48.22\%$ global accuracy, ahead of the strongest competitor DFML ($47.29\%$) and well above the uniform average ($41.53\%$) and Uncertainty KD ($42.06\%$). Two patterns stand out: the margin over uniform averaging widens sharply relative to CIFAR-10, from about one point to nearly seven, and DFML now overtakes the uniform and uncertainty-based variants that outscored it on CIFAR-10. Both follow from the same cause. With $100$ classes and far fewer samples per class, per-class beliefs are noisier and locally-rare classes more numerous, the regime in which tying a teacher's weight to its reliability pays off.

\paragraph{Consistency on a Medical Benchmark.}
PathMNIST instantiates the clinical scenario of Section~\ref{sec:intro}: sensitive slides that sites cannot share, held at institutions running widely differing hardware. Its columns in Table~\ref{tab:main} show CRAD again attaining the best global accuracy ($89.44\%$, against $87.59\%$ for the uniform average) and the best local accuracy ($88.14\%$). That the gain transfers from natural images to histopathology indicates that reliability-aware aggregation is not specific to one data domain.

\begin{table}[tb]
\centering
\caption{Cost on CIFAR-10 ($N=10$), measured during the runs of Table~\ref{tab:main} with every method on the same machine (specified in Appendix~\ref{sec:supp_compute}). Payload is what one client transmits per round, Time the wall-clock duration of a full communication round, and GPU Mem.\ its peak allocated GPU memory. ``Proxy'' marks methods additionally requiring public or synthetic data.}
\label{tab:cost}
\footnotesize
\setlength{\tabcolsep}{8.2pt}
\resizebox{\columnwidth}{!}{%
\begin{tabular}{lcccc}
\toprule
Method & Payload & Time & GPU Mem. & Proxy \\
       & (MB) & (s) & (GB) &  \\
\midrule
\multicolumn{5}{l}{\textit{Competing heterogeneous-FL methods}}\\
FedMD~\cite{Li2019FedMD}          & {0.38} & {37.6} & {1.38} & yes \\
FedGD~\cite{zhang2023target}      & {0.04} & {26.2} & {3.33} & yes \\
MSFKD~\cite{Wang2023Knowledge}    & {0.04} & {25.1} & {1.57} & yes \\
DFML~\cite{Khalil2024DFML}            & {38.9} & {53.8} & {1.20} & no \\
FedMKD~\cite{Lin2025FedMKD}       & {194.3} & {46.1} & {1.56} & no \\
\midrule
\multicolumn{5}{l}{\textit{Our framework, varying teacher aggregation}}\\
Uniform KD                        & {194.3} & {20.6} & {0.60} & no \\
Uncertainty KD                    & {194.3} & {20.0} & {0.60} & no \\
\rowcolor{gray!20} \textbf{CRAD (Ours)} & {194.3} & {21.3} & {0.60} & no \\
\bottomrule
\end{tabular}}
\end{table}

\subsection{Analyses}

\paragraph{Per-Class Gains.}
\begin{figure}
    \centering
    \includegraphics[width=\linewidth]{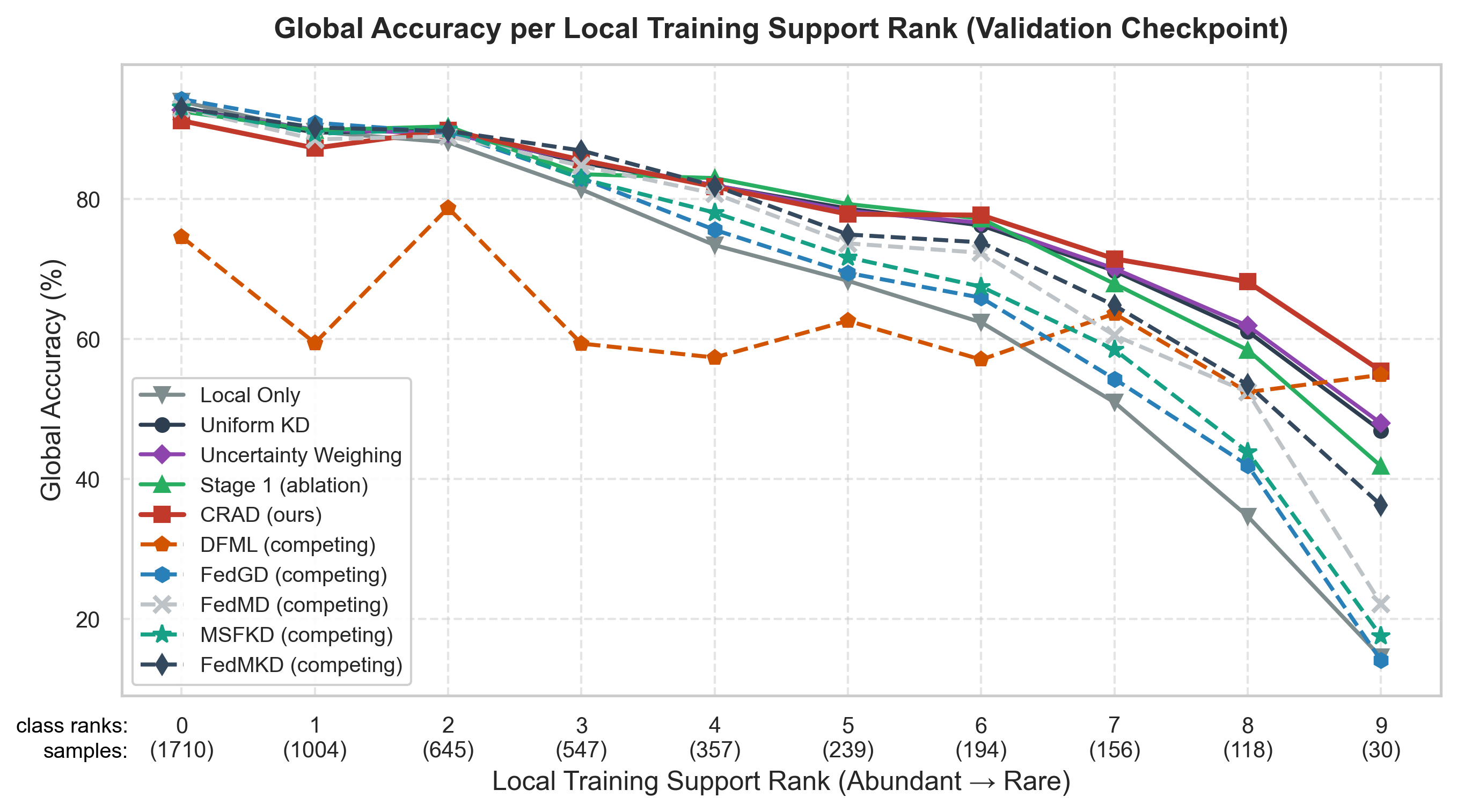}
    \caption{Per-class global accuracy on CIFAR-10 with classes ordered by local training support (shown in parentheses, abundant to rare), for all compared methods. The methods agree more on well-supported classes and separate on the rare ones, where CRAD degrades the least.}
    \label{fig:per_class_all}
\end{figure}
Figure~\ref{fig:per_class_all} breaks global accuracy down by class, ordered from abundant to rare local support. On a class that a client holds few samples of, the uniform average dilutes the few competent teachers among a majority of uninformed ones, and a teacher can be confidently wrong on a class it has never seen. Tying the weight to the teacher's reliability down-weights the uninformed ones. The figure bears this out: the methods are nearly indistinguishable where support is plentiful and separate as it dwindles. On the rarest class (30 local samples), CRAD retains about $56\%$ accuracy against about $47\%$ for the other aggregation variants, and it leads every competing method on the five rarest classes. 
The Stage-1 ablation, CRAD's agreement filter without the precision weight, does no better than the uniform average on these rare classes, so the lift comes mainly from precision-weighting the survivors, not from the filter alone.

\begin{figure}[t]
    \centering
    \includegraphics[width=\linewidth]{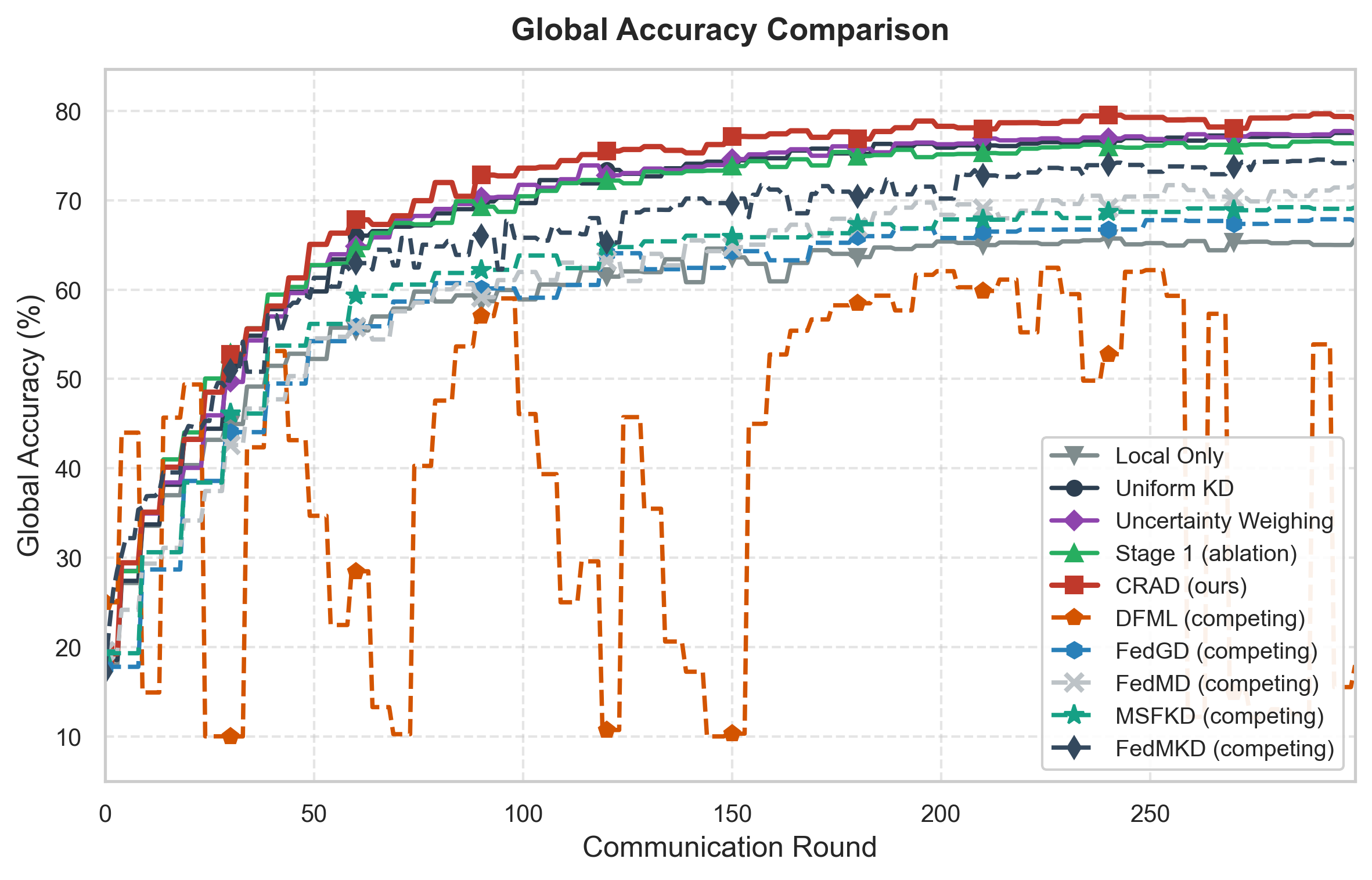}
    \caption{Global accuracy vs. communication round for all compared methods on CIFAR-10. CRAD (red) stays above every other method from the early rounds onward and converges stably, whereas DFML swings sharply from round to round.}
    \label{fig:cifar10_curves}
\end{figure}

\paragraph{Convergence and Stability.}
Figure~\ref{fig:cifar10_curves} plots global accuracy over communication rounds. CRAD sits above every baseline, including the Stage-1 ablation, from the early rounds onward and converges smoothly, whereas DFML's curve swings sharply round to round, so the round selected on validation sits far above its typical accuracy.

\paragraph{Communication, Compute, and Memory Cost.}
Table~\ref{tab:cost} shows that cost is set by the regime, not by the aggregation rule. Proxy-based methods communicate only predictions over the transfer set and are cheap, but require that set to exist, whereas every snapshot-based decentralized method exchanges full snapshots. Within that regime, CRAD's own overhead is the $2C$-scalar statistics vector, under 1\,KB, and it completes a round faster and at lower peak memory than every competing method. Round time and peak memory are both dominated by holding and running the $N{-}1$ peer teachers, hence nearly identical across our three aggregation variants. The higher payload is the price of requiring neither public data nor a server.

\paragraph{Client-Number Ablation.}
    To test scalability under sparse participation, we grow the pool to $N\in\{20,50,100\}$ clients while activating 10 per round. As shown in Table~\ref{tab:client_ablation}, CRAD improves both global and local accuracy over Uniform KD at every pool size, and its relative advantage widens as the pool grows, from a factor of $1.2$ on global accuracy at $N{=}20$ to $2.0$ at $N{=}100$. This suggests that CRAD is particularly effective under sparse participation, where teacher quality varies more widely.

\begin{table}[tb]
\centering
\caption{Client-number ablation on CIFAR-10 with partial participation. Each run uses $R=300$ rounds and activates 10 clients per round. Accuracies are averaged over all clients.}
\label{tab:client_ablation}
\setlength{\tabcolsep}{5pt}
\footnotesize
\resizebox{\columnwidth}{!}{%
\begin{tabular}{lcccccc}
\toprule
& \multicolumn{2}{c}{$N{=}20$} & \multicolumn{2}{c}{$N{=}50$} & \multicolumn{2}{c}{$N{=}100$} \\
\cmidrule(lr){2-3}\cmidrule(lr){4-5}\cmidrule(lr){6-7}
Method & Global & Local & Global & Local & Global & Local \\
\midrule
Uniform KD & 53.77 & 68.89 & 22.81 & 54.02 & 11.01 & 40.83 \\
\textbf{CRAD (Ours)} & \textbf{64.18} & \textbf{72.58} & \textbf{29.94} & \textbf{58.16} & \textbf{22.19} & \textbf{55.83} \\
\bottomrule
\end{tabular}}
\end{table}

\begin{table}[tb]
\centering
\caption{Ablation of teacher-weighting rules on CIFAR-100. All variants use the same CRAD framework and differ only in the aggregation weight. Acc.: accuracy-only; Sup.: support-only; SM: $\mathrm{softmax}(-\sigma^2)$; LCB: $\max(0,\tilde a-\sigma)$; $\mathrm{CE}$: cross-entropy.}
\label{tab:cifar100_weight_ablation}
\footnotesize
\resizebox{\columnwidth}{!}{%
\begin{tabular}{lccccccc}
\toprule
 & $1/\sigma$ & Acc. & Sup. & SM & LCB & $1/\mathrm{CE}$ & \textbf{CRAD} \\
\midrule
Global Acc. & 46.89 & 47.22 & 47.25 & 46.71 & 47.02 & 47.17 & \textbf{48.22} \\
Local Acc.  & 57.98 & 58.56 & 58.26 & 58.45 & 58.54 & 58.53 & \textbf{59.87} \\
\bottomrule
\end{tabular}}
\end{table}

\paragraph{Weighting Rule.}
Table~\ref{tab:cifar100_weight_ablation} swaps the precision weight for six alternatives on CIFAR-100, holding the rest of CRAD fixed. Precision weighting is best on both metrics. Accuracy alone and support alone each capture part of the gain, but combining them via precision weighting beats either. Weighting by $1/\sigma$ rather than $1/\sigma^2$ is the weakest variant on local accuracy, consistent with precision being the scale on which evidence adds (Sec.~\ref{sec:crad}).

\section{Limitations and Future Work}
\label{sec:limitations}
CRAD requires sharing a per-class statistics vector, which reveals each client's label proportions through the training counts $n_{\mathrm{train},j}^c$ and the validation counts $n_j^c$. These are fixed across rounds, so a single noised release suffices and its privacy cost does not compose over rounds. Each sample lies in one split, so the counts jointly have $L_1$-sensitivity one under record add/remove adjacency, and Laplace noise of scale $1/\varepsilon$ on all of them would give $\varepsilon$-differential privacy. At $\varepsilon=1$, noising only the validation counts costs $0.82$ points of global accuracy on CIFAR-100 ($48.22\%$ vs.\ $47.40\%$). Noising both the training and validation counts, which gives $\varepsilon$-differential privacy for the shared counts, costs $1.21$ points ($47.01\%$), within $0.3$ points of the strongest competing method, DFML ($47.29\%$), and ahead of all others in Table~\ref{tab:main}. End-to-end differential privacy covering the snapshots remains open and is left to future work. 

\section{Conclusion}

We presented a decentralized, server-free, public-data-free framework in which architecturally heterogeneous clients collaborate purely through peer distillation, and proposed CRAD, a per-class aggregation that filters teachers by consensus agreement and weights the survivors by the precision of their class-accuracy estimates. Across CIFAR-10/100 and PathMNIST under heterogeneous architectures and severe non-IID skew, CRAD attains the best global accuracy.
\paragraph{Note on this version.} This version addresses a discrepancy between the method description (Eq.~\ref{eq:agreement_score} and the support filter) in the WACV 2027 proceedings version and the code used for the experiments, and adds one privacy experiment. All results reported in the proceedings are unchanged.

{
    \small
    \bibliographystyle{ieeenat_fullname}
    \bibliography{main}
}

\appendix

This appendix is organized as follows: Appendix~\ref{sec:supp_hyperparams} specifies optimization settings, hyperparameters, the support pre-filter, evaluation conventions, and the compute environment. Appendix~\ref{sec:supp_qualitative} provides a qualitative view of how CRAD forms its per-class
distillation target on individual test images, including a case where the
mechanism fails. It uses the same CIFAR-10 run as Table~\ref{tab:main},
read at the same best-validation checkpoint.

\section{Extended Implementation and Hyperparameter Details}
\label{sec:supp_hyperparams}

\subsection{Hyperparameter Summary}

Table~\ref{tab:hyperparameters} summarizes the exact hyperparameter values across all three benchmark datasets (CIFAR-10, CIFAR-100, and PathMNIST).

\begin{table*}[t]
\centering
\small
\caption{\textbf{Comprehensive Hyperparameter Configuration.} Parameters used in the experimental benchmark runs for CRAD across CIFAR-10, CIFAR-100, and PathMNIST.}
\label{tab:hyperparameters}
\vspace{4pt}
\begin{tabular}{llccc}
\toprule
\textbf{Category} & \textbf{Hyperparameter} & \textbf{CIFAR-10} & \textbf{CIFAR-100} & \textbf{PathMNIST} \\
\midrule
\multirow{8}{*}{\textbf{Local Training}}
& Local Optimizer & SGD & SGD & SGD \\
& Initial Learning Rate & 0.1 & 0.1 & 0.1 \\
& Momentum & 0.9 & 0.9 & 0.9 \\
& Weight Decay & $5 \times 10^{-4}$ & $5 \times 10^{-4}$ & $5 \times 10^{-4}$ \\
& Learning Rate Decay ($\times 0.99$/round) & Exponential & Exponential & Exponential \\
& Local Batch Size & 64 & 64 & 64 \\
& Communication Rounds ($R$) & 300 & 300 & 300 \\
& Local Epochs per Round & 1 & 1 & 1 \\
\midrule
\multirow{5}{*}{\textbf{Distillation (CRAD)}}
& Distillation Temperature ($T$) & 4.0 & 4.0 & 4.0 \\
& Distillation Weight ($\lambda$) & 0.7 & 0.7 & 0.7 \\
& Support Threshold ($n_{\min}^c$) & Top 50\% & Top 50\% & Top 80\% \\
& Agreement Filter Cutoff ($\tau_c$) & Median Deviation & Median Deviation & Median Deviation \\
& Variance Numerical Guard ($\epsilon$) & $10^{-6}$ & $10^{-6}$ & $10^{-6}$ \\
\midrule
\multirow{3}{*}{\textbf{Splits \& Checkpoints}}
& Local Evaluation Split Protocol & 50/50 val/test & 50/50 val/test & 50/50 val/test \\
& Local Validation Samples per Client & $\approx 500$ & $\approx 530$ & $\approx 361$ \\
& Checkpoint Selection Criterion & Mean Local Val. & Mean Local Val. & Mean Local Val. \\
\bottomrule
\end{tabular}
\end{table*}

\subsection{Validation Splitting}
Each client's partitioned evaluation set is split 50/50 into a dedicated local validation split and an untouched local test split. The validation split is used only for computing local empirical validation accuracies $a_j^c$, validation class sample counts $n_j^c$, and checkpoint selection.

\subsection{Support Pre-Filtering}
Before the Stage-1 agreement filter and Stage-2 precision weighting, peers lacking sufficient local training support for class $c$ are pruned by the support cutoff $n_{\min}^c$. On CIFAR-10 and CIFAR-100, $n_{\min}^c$ is set to the peer median, retaining the top 50\% of peers with the highest local training class support. On PathMNIST, which exhibits higher class density, $n_{\min}^c$ is set to the 20th percentile, retaining the top 80\% of peers by local training class support.

\subsection{Checkpoint Selection and Evaluation Metrics}
\begin{itemize}
    \item \textbf{Validation-Based Checkpoint Selection:} Following standard federated learning protocol, the checkpoint is selected using validation performance rather than test performance. At each round, each client evaluates its model on its local validation split. The round achieving the highest unweighted mean validation accuracy across all $N=10$ clients is selected as the optimal checkpoint: Round 228 for CIFAR-10, Round 277 for CIFAR-100, and Round 285 for PathMNIST.
    \item \textbf{Global Accuracy:} The model snapshots saved at the selected optimal round are evaluated individually on the official benchmark test set (10,000 samples for CIFAR-10/CIFAR-100; 7,180 samples for PathMNIST). The reported accuracy is the unweighted macro-average over all $N$ client models, rather than an ensemble evaluation.
\end{itemize}

\subsection{Compute Environment}
\label{sec:supp_compute}
Every experiment used a single NVIDIA A100-SXM4-80GB GPU on the Cheaha high-performance computing cluster at the University of Alabama at Birmingham, with 6 AMD EPYC CPU cores and 60\,GB of host memory allocated per job. The software stack was PyTorch 2.6.0, CUDA 12.4, and Python 3.11.4 on Linux. The timing and memory figures in Table~\ref{tab:cost} come from the same runs as the accuracy figures in Table~\ref{tab:main}, all executed under this configuration. GPU Mem.\ reports the peak GPU memory allocated during each run.

\section{Qualitative Mechanism on Individual Images}
\label{sec:supp_qualitative}

\begin{figure*}[tp]
    \centering
    \includegraphics[width=0.85\textwidth]{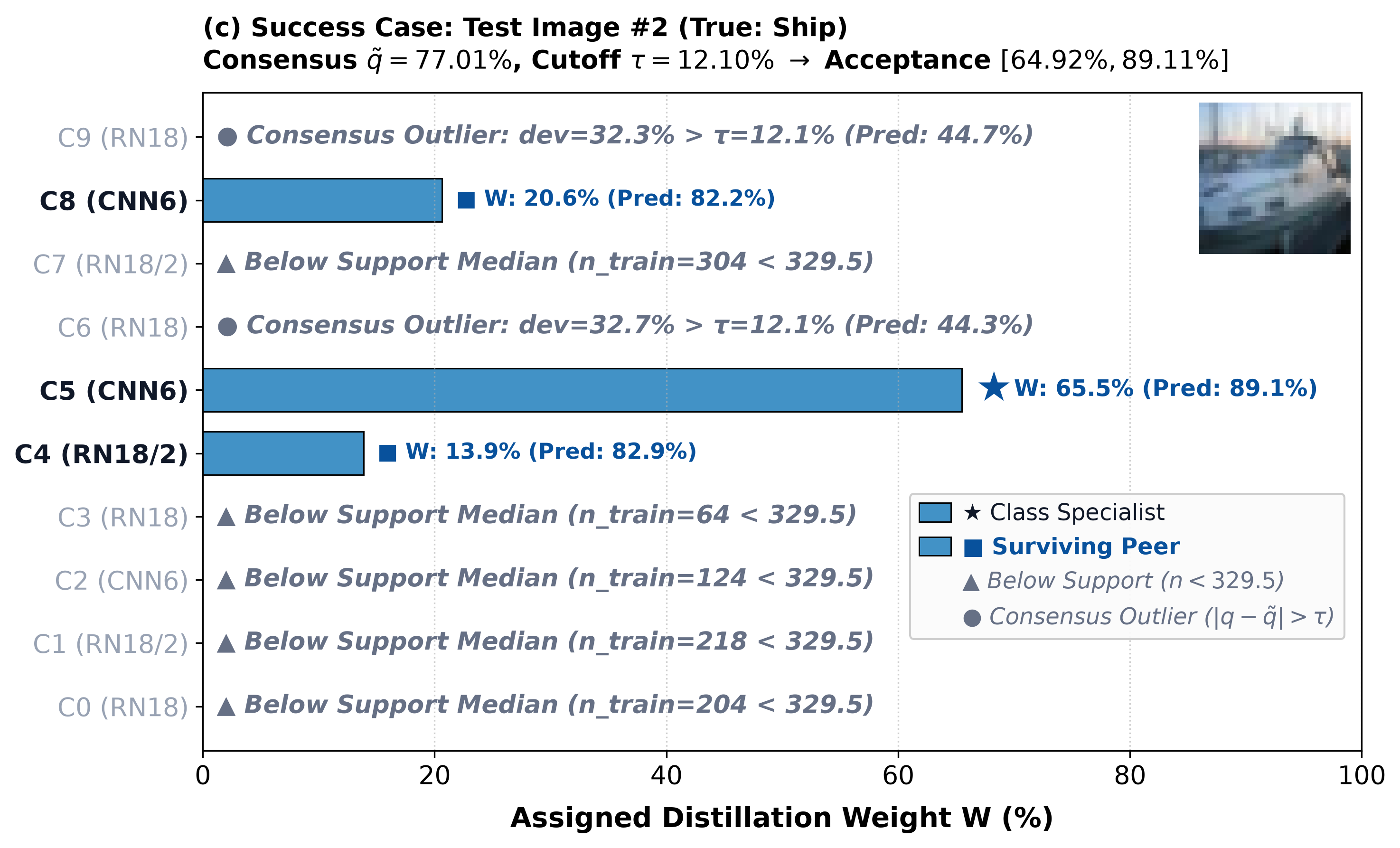}\\[6pt]
    \includegraphics[width=0.85\textwidth]{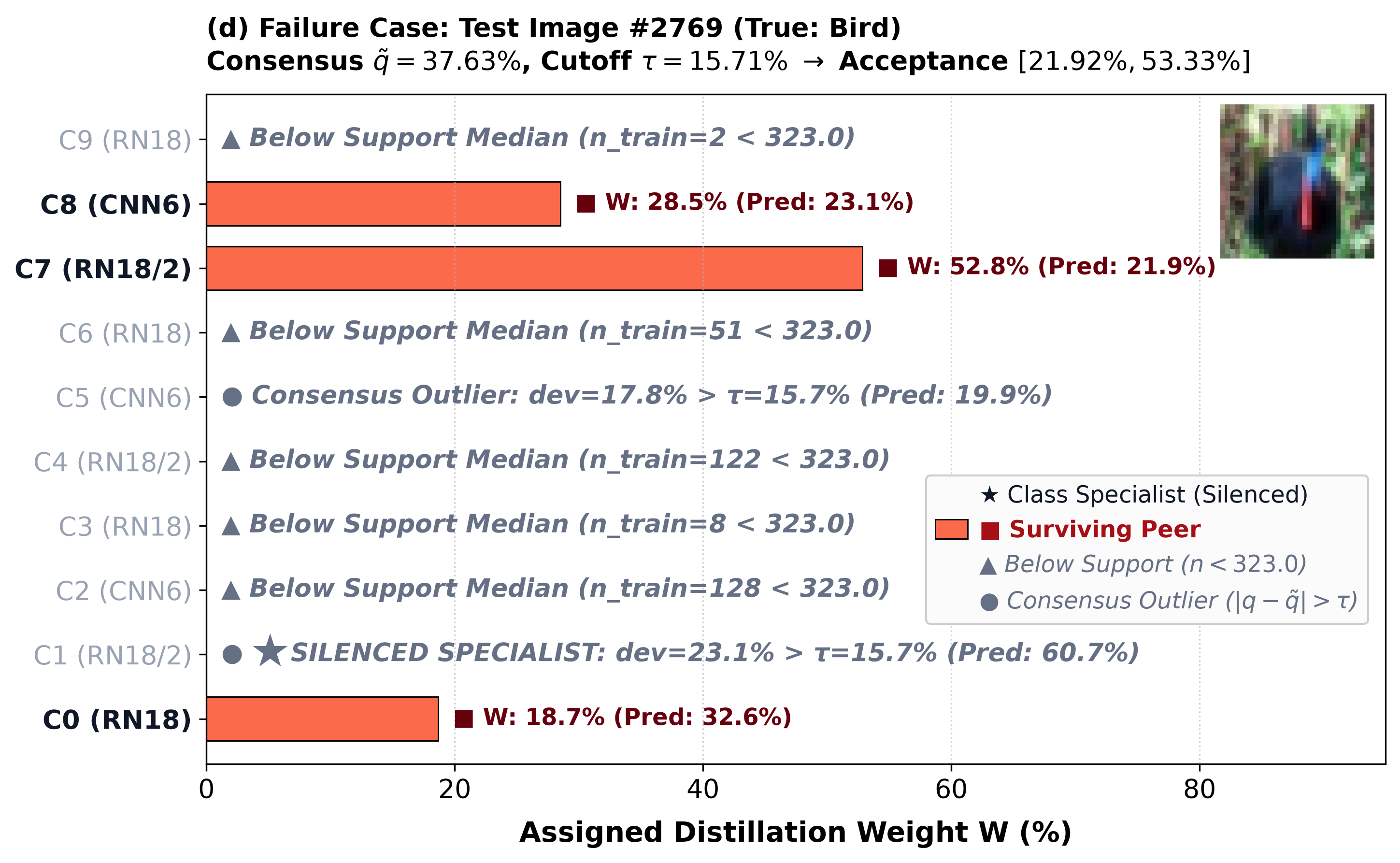}
    \caption{CRAD's two-stage aggregation traced on two individual CIFAR-10 test images. $W$ is the assigned distillation weight after per-class normalization, the parenthesized value is the client's probability on the true class, and $\text{dev} = |q_j^c - \tilde{q}^c|$ denotes the client's absolute deviation from the preliminary consensus $\tilde{q}^c$. 
    \textbf{(c) Success case, a ship image:} peers below the median training support threshold ($n < 329.5$) are dropped as uninformed, two more fall outside the acceptance band $\tilde{q}^c \pm \tau_c$ and are dropped as consensus outliers, and the surviving well-supported specialist (C5, $n = 1{,}657$) receives $65.5\%$ of the weight, raising the aggregated target on the true class to $86.8\%$ from a preliminary consensus of $77.0\%$ (vs. $58.9\%$ for uniform KD). \textbf{(d) Failure case, a bird image:} the best-evidenced teacher (C1, $n = 1{,}798$, and correct at $60.7\%$) deviates from a low consensus formed by less informed peers and is removed by Stage~1 (agreement filter), so its knowledge never reaches Stage~2 (precision weighting) and the target stays at $24.2\%$.} \label{fig:cases} \end{figure*}

The per-class breakdown in Fig.~\ref{fig:per_class_all} shows which classes reliability-aware
aggregation helps on, averaged over clients and test images. They do not show
how the two stages act on any single prediction. Figure~\ref{fig:cases}
resolves one success case and one failure case down to the individual teacher.

\begin{figure*}[tp]
    \centering
    \includegraphics[width=0.85\textwidth]{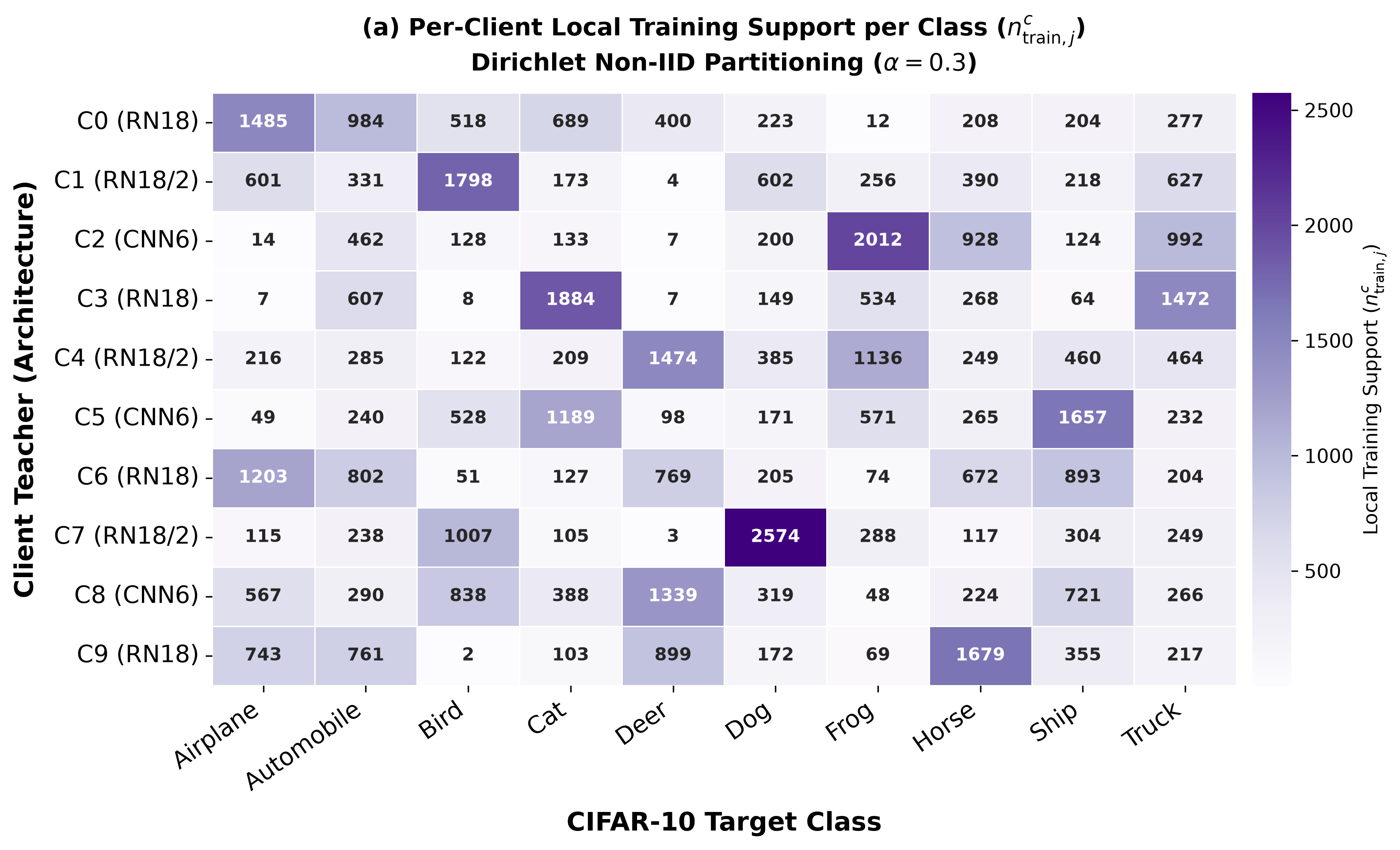}\\[6pt]
    \includegraphics[width=0.85\textwidth]{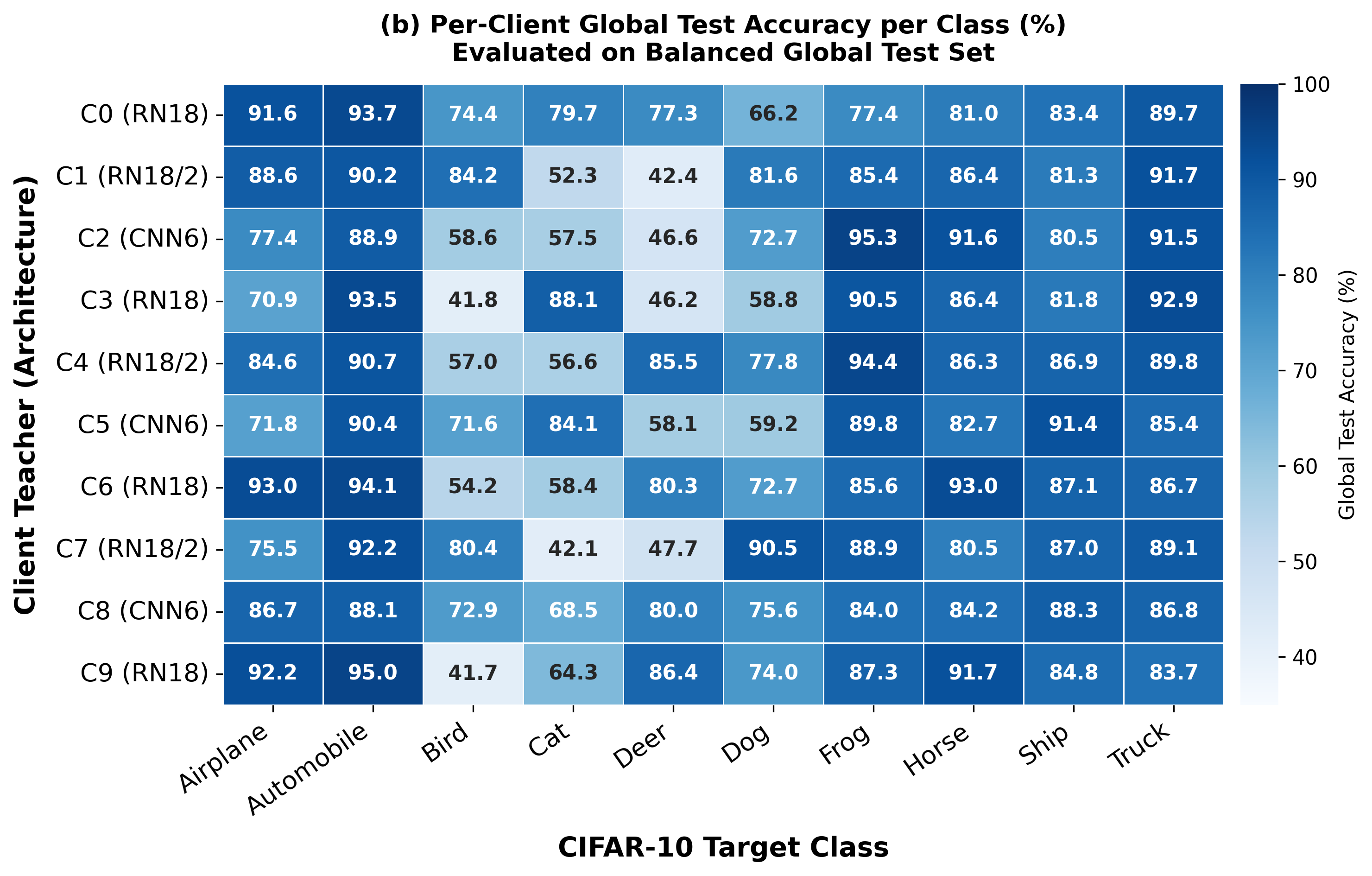}
    \caption{Context that drives the weighting, on CIFAR-10.
    \textbf{(a)} Per-client local training support $n_{\mathrm{train},j}^c$ under the Dirichlet
    ($\alpha{=}0.3$) partition, governing the support pre-filter (each client's held-out validation split contains $10\%$ of these counts, supplying $n_j^c$ to Eq.~6).
    \textbf{(b)} Per-client accuracy on the balanced global test set, by class.
    The two are related but not identical, which is the reason CRAD weights on
    both accuracy and the evidence behind it rather than on either alone.}
    \label{fig:context}
\end{figure*}

\paragraph{Setup.}
Figure~\ref{fig:context} gives the context that drives the weights. Panel~(a) is the
per-client local training support $n_{\mathrm{train},j}^c$ under the Dirichlet
($\alpha{=}0.3$) partition, the count that drives the support pre-filter. Each client's validation split holds about $10\%$ of these counts, and those validation counts $n_j^c$ enter the variance of
Eq.~\ref{eq:variance} through Agresti--Coull smoothing. Panel~(b) is the
per-client accuracy on the balanced global test set, broken down by class.
The two are related but not identical, which is the reason CRAD weights on
both accuracy and the evidence behind it rather than on either alone.

Clients are labeled by architecture, with RN18, RN18/2, and CNN6 denoting ResNet-18, ResNet-18-Half, and the plain CNN of Sec.~\ref{sec:experiments}.

The two panels of Fig.~\ref{fig:cases} trace one test image each. For every client we report the
assigned distillation weight $W$, that is the precision weight of
Eq.~\ref{eq:precision_weight} after per-class normalization, and in parentheses the
probability that client assigns to the true class. Clients are marked in one
of three states: retained and weighted, dropped for insufficient class support,
or dropped as a consensus outlier. The acceptance band printed above each panel
is $\tilde{q}^c \pm \tau_c$, where $\tilde{q}^c$ is the preliminary consensus and
$\tau_c$ the median absolute deviation from it, so by construction roughly half
of the supported teachers survive Stage~1.

\paragraph{Success case (Fig.~\ref{fig:cases}c).}
On a ship image, five clients hold fewer than the peer median of $329.5$ local training samples of the class and are dropped as uninformed before any weighting occurs. Of the remaining five, two fall outside the acceptance band and are discarded as consensus outliers. Of the three survivors, the client holding $1{,}657$ ship training samples receives $65.5\%$ of the total weight and assigns $89.1\%$ to the correct class, while the two lower-support survivors split the remainder. The aggregated target is therefore dominated by the one teacher that is both well-evidenced and confident, and it rises to $86.8\%$ on the true class, well above both the preliminary consensus of $77.0\%$ and the $58.9\%$ that a uniform average would have produced.

\paragraph{Failure case (Fig.~\ref{fig:cases}d).} The same mechanism can suppress the teacher it should trust. On a bird image the peer consensus is low ($37.6\%$), with a median deviation of $\tau_c = 15.7\%$. Client~1 holds $1{,}798$ training samples of the class, by a wide margin the best-evidenced teacher, and assigns $60.7\%$ to the correct class. It is also the most accurate teacher on that class, at $84.2\%$ in panel~(b), so the filter is discarding the one teacher the weighting was designed to favor. Because that estimate lies $23.1$ points from a consensus formed by less informed peers, Stage~1 removes it before Stage~2 can weight it. The surviving teachers all sit near or below the low consensus value, and the aggregated target ends at $24.2\%$, below even the consensus it started at.

This is the cost of consensus filtering under the assumption stated in
Sec.~\ref{sec:crad}, that unreliable teachers are in the minority for a class. Where that
assumption holds the filter is protective, and where a well-supported specialist
disagrees with a less informed majority it can remove useful signal. Letting
support temper the consensus filter, rather than applying the two steps in
sequence, is a natural refinement that we leave to future work.

\balance   
\end{document}